# Leveraging Expert Models for Training Deep Neural Networks in Scarce Data Domains: Application to Offline Handwritten Signature Verification

Dimitrios Tsourounis [a,*], Ilias Theodorakopoulos [b], Elias N. Zois [c] and George Economou [a]

[a] *Department of Physics, University of Patras, 26504, Rio, Greece*
[b] *Department of Electrical and Computer Engineering, Democritus University of Thrace, Greece*
[c] *Department of Electrical and Electronic Engineering, University of West Attica, Greece*

## Abstract

This paper introduces a novel approach to leverage the knowledge of existing expert models for training new Convolutional Neural Networks, on domains where task-specific data are limited or unavailable. The presented scheme is applied in offline handwritten signature verification (OffSV) which, akin to other biometric applications, suffers from inherent data limitations due to regulatory restrictions. The proposed Student-Teacher (S-T) configuration utilizes feature-based knowledge distillation (FKD), combining graph-based similarity for local activations with global similarity measures to supervise student's training, using only handwritten text data. Remarkably, the models trained using this technique exhibit comparable, if not superior, performance to the teacher model across three popular signature datasets. More importantly, these results are attained without employing any signatures during the feature extraction training process. This study demonstrates the efficacy of leveraging existing expert models to overcome data scarcity challenges in OffSV and potentially other related domains.

* Corresponding Author. Physics Dept., University of Patras, University Campus, 26504 Rio, Greece

Email addresses: dtsourounis@upatras.gr (D. Tsourounis), iltheodo@ee.duth.gr (I. Theodorakopoulos), ezois@uniwa.gr (E. N. Zois), economou@upatras.gr (G. Economou)

# 1   Introduction

A signature is a handwritten symbol put on a document, a piece of paper or other material that allows authenticating someone's identity and consent. Although there are findings from Sumerian civilization using words and symbols to denote identity, the legal usage of signatures grew in Europe during the 16$^{th}$ century with the voted Act of Statute of Frauds, which stipulated that contracts must exist in writing and bear a signature, and became a standard form around the world since it adopted in colonial America (by the time J. Hancock placed his signature in the United States Declaration of Independence) for validating agreements. Nowadays, the determination of authenticity through signatures is employed in many sectors to ensure the security of financial and legal documents ranging from bank and compliance forms to contracts and mail ballots. While manual signature comparison seems like an ineffective way to handle the masses of documents that need to be checked in a small amount of time, automatic handwritten signature verification systems (HSV) are pivotal to reduce fraud. A HSV system authenticates the person's identity on the basis of the claimed identity, meaning that detects authenticity (i.e., the questioned signature owns to the claimed writer and thus it is genuine) or whether the signature has been provided from anyone else and thus it is forgery.

According to the acquisition conditions, signature verification is divided into offline (static) and online (dynamic). Offline Signature Verification (OffSV) analyzes only the shape (visual information) of the signature after the writing process, typically using a digitized version of the signing document, while Online Signature Verification (OnSV) requires a digitizing device (such as electronic tablets) and collects additional information during signing, like pen inclination, pressure, spatial coordinates, etc. (Plamondon & Lorette, //). Hence, in the first case, the signature is represented as a static (digital) image, making the OffSV a very challenging problem due to the luck of dynamic measurements. On the other, it is more practical than OnSV because the latter involves rushed, abnormal, and device-depending operations that might decrease the trustworthiness of the produced signatures (Impedovo & Pirlo, 2008).

Depending on the signature verification design plan, there are two main approaches: writer-dependent (WD) and writer-independent (WI). WD methods build one model per user and WI approach uses one single (global) model for all users. Given that, a typical HSV system comprises of three main stages: data acquisition and preprocessing (for noise removal), feature extraction, and classification (decision on authenticity), some stages could follow the WD approach and others the WI approach, leading to hybrid systems. Thus, a characterization of a HSV system as WD or WI is often not easy, especially when complex deep learning schemes are utilized, which commonly consist of WI preprocessing and WI feature extraction stages along with WD classification (Diaz et al., 2019). The WI methods usually take advantage of the increased number of training samples by generating pairs between signatures (generally a similar pair includes signatures of the same writer while a dissimilar pair combines genuine and forgery signatures) and embed the feature representations to a dissimilarity space for obtaining the final decision. The WD approaches rely on the signatures of each writer to create a custom model dedicated to each writer and thus, they turn to be more restricted but at the same time they could provide more efficient results due to tailoring on each individual (Hafemann et al., 2017b).

Three main challenges should be addressed by an HSV system. First, the high intra-class variability of handwritten signatures, even though they are provided by the same user. This results from the complex mechanisms underlying handwriting that depend on the psychophysical state of the signer and the conditions under which the signature apposition process occurs (Impedovo & Pirlo, 2008). Secondly, the partial knowledge during the deployment of the system since only the genuine signatures of the enrolled users are available whilst the system should be able to accept genuine and reject forgery signatures. This task is even more challenging when considering the presence of different types of forgery signatures: i) the random forgeries which are generated without access to the original signature, ii) the simple or unskilled forgeries where the forger has information about the shape of genuine but not allowed much practice during falsification, and iii) the skilled forgeries, as the most difficult case, where the impostor attempts to carefully imitate the original signature with no constraints (Pal et al., 2011). The last type presents also low inter-class variability since skilled forgeries resemble genuine signatures, casting the verification task harder. Thirdly, the limited number of available samples for each user, since only a few signatures are supplied from each one user in a realistic scenario. Therefore, while many users could be enrolled, the system needs to perform well for any new user from whom a small number of signatures is provided every single time.

While the above challenges are inherent on the OffSV problem, an additional extrinsic limitation is introduced from the absence of large offline datasets (Diaz et al., 2019). Until recently, the GPDS-960 corpus offline database (Vargas et al., 2007), with more than half a thousand writers having 24 genuine along with 30 forgeries signatures per writer, allowed the training of deep models into the similar task of writer identification (Hafemann et al., 2017b). Even though these CNN models are not specialized to the task of signature verification, the large size of GPDS-960 dataset enabled CNNs to be good universal functions for producing image-level feature descriptors for signature images, surpassing the expressiveness of hand-crafted features (Hameed et al., 2021). Unfortunately, this dataset, is no longer available due to the General Data Protection Regulation (EU) 2016/679 ("GDPR"), thus hindering the efforts of research community to investigate new models and design elaborate methods that require more training data.

Motivated by the data-intensive nature of CNNs' training, many OffSV systems pursue designing methodologies to address the lack of adequate signature training data. These approaches follow two main directions, the generation of synthetic signature images using geometrical transformations (Diaz et al., 2017; Parmar et al., 2020) or generative learning models (Jiang et al., 2022; Natarajan et al., 2021; Yapıcı et al., 2021; Yonekura & Guedes, 2021) and the utilization of images from a relative domain such as the handwritten text documents (Mersa et al., 2019; Tsourounis et al., 2022). For completeness, there are also developed feature space augmentation methods that artificially populate samples for improving the classifier' performance, yet they rely on feature vector representations and do not create signature images for training (Arab et al., 2023; Maruyama et al., 2021; Zois et al., 2023). Finally, considering the fully synthetic nature of generated signatures and the contingent unreal identities, the signature duplications would be prone to diverge from the realistic intra-subject variability criterion and thus, their use as the training images of an end-to-end deep learning model could be problematic, requiring special manipulations for their beneficial usage output (Viana et al., 2022, 2023).

Despite the capabilities of the above approaches to cope with signature verification problems with small sample size, these methods ignored the knowledge of older benchmark models in the OffSV task. In situations where an effective CNN model is available, a popular approach is to transfer knowledge from this (expert) model to facilitate learning of another (new) model. This forms the case of Knowledge Distillation (KD), where the knowledge is transferred between the models that assume the role of teacher (expert) and student (new) (Wang & Yoon, 2022). When the teacher is pre-trained and fixed during training, it is called offline KD. Additionally, the condition when an effective teacher model exists but there is no access to its training data constitutes the data-free KD, where the distillation process uses only external or artificial data to perform the knowledge transfer from the teacher to the student model (Gou et al., 2021). Finally, another branch which is relevant to this work is Feature-based KD (FKD), which involves distilling knowledge from the intermediate layers of the teacher model in order for the student model to learn feature representations that are a good approximation of the teacher's intermediate representations.

To the best of our knowledge, we consider this work to be the first one that introduces the data-free KD approach into the OffSV domain. Here we propose a novel KD method to transfer the knowledge from a teacher CNN into a new CNN student model with different architecture. This allows the new model to leverage the knowledge learned by the teacher model, even though the original training data are not available anymore. Furthermore, the new model is able to achieve improved performance on the OffSV task compared to the teacher. The KD scheme consists of 1) the teacher CNN supervising the training process, 2) the training data used to transfer the knowledge, and 3) the KD method that defines knowledge features, distillation loss, strategy, and connections. The ultimate goal is to express the learned information inasmuch as it is helpful for building up a well-performing student CNN. Therefore, to address OffSV using offline data-free Feature-based KD, the above components are realized as follows:

1) An appropriate teacher is one of the benchmark CNN models in the field, such as SigNet (Hafemann et al., 2017a) which is trained with the genuine signature images from 531 writers using GPDS-960 corpus and the trained model is publicly available[1]. In this occasion, the teacher model can provide valuable feature representations for any input image, but not a meaningful classification response, since the training classes are person IDs which are irrelevant outside the specific identification task.
2) The data which act as information carriers for the distillation, can be either synthetic or external. In our work, we opt to utilize images of handwritten text because they possess a similar structure to signatures (thin pen strokes on a piece of paper) and most importantly, there is an abundance of data available from public sources. The option of using synthetically generated signatures was dismissed in light of evidence indicating that the currently available synthetic signature datasets can deteriorate the effectiveness of OffSV systems if used to train the feature extractor model (Viana et al., 2022, 2023; Yapıcı et al., 2021).

---

[1] https://github. com/luizgh/sigver/tree/master/sigver/featurelearning/models

3) The Feature-based Knowledge Distillation (FKD) is applied for guiding the activations at the intermediate layers of teacher and student models. Here, we utilize computationally efficient loss functions aiming to transfer the geometry of activations from intermediate layers of the teacher CNN to the activations of the student CNN model at matching spatial resolutions. The employed loss functions emerge from manifold-manifold distance functions, formulating the problem of FKD as a problem of learning similar manifolds of local activations in corresponding layers of teacher and student models. Furthermore, the training of the student model incorporates KD attained by an additional regularization loss that is based on the global feature, generated at the penultimate layers of the teacher and student models respectively. Under this direction an efficient loss function is designed to fit with the KD scope, inspired from the Self-Supervised Learning method of Barlow Twins (Zbontar et al., 2021). Ultimately, the proposed KD method utilizes both geometric FKD and global FKD, thus integrating local information via manifold-to-manifold comparison as well as global information via metrics that range from typical temperature-scaled cross entropy to KD-oriented cross-correlation losses.

4) The requirements for the student CNN model architecture, utilized in the FKD scheme are: (i) matching of intermediate activations for at least some of spatial resolutions and (ii) for the global feature to share equal dimensions with that of the teacher model. The popular ResNet-18 CNN was selected as the student architecture, given its efficiency and modern topology (He et al., 2015).

The training of a feature extraction model for OffSV is a learning task different than the main verification task, since the identity and data of the users involved in the operational phase are not always available during the model's training. Following feature extraction, the decision stage analyses the feature representation of a signature image and decides upon its validity. Since the goal of this work is to demonstrate the value of the proposed FKD in designing an efficient OffSV feature extractor, at the final decision stage we follow the most straightforward WD approach, using WD Support Vector Machine classifiers to evaluate our method at the operational phase. Results indicate that our system achieves top-tier performance on three popular Latin offline signature datasets without requiring any signature images during Student-Teacher training. The verification error is in par with state-of-the-art models trained with thousands of signature images, obtained by only exploiting knowledge via the proposed FKD scheme. Also, the training of the OffSV system does not require any skilled forgery signatures because the final decision stage with the WD classifiers uses only genuine signatures and particularly, a few signatures of the writer along with some signatures of other writers, also known as random forgeries.

The contributions of the proposed work for OffSV could be summarized as follows:

- We demonstrate that FKD enables the efficient training of any new architectures that inherit the prior knowledge of benchmark models whose training data are unavailable, using external data of similar nature.

- The knowledge transfer between expert CNN model and new CNN model is accomplished without the use of signature images, employing only handwritten text data processed using a specialized yet simple pre-processing scheme.
- We propose a method for KD that combines information from both local features' geometry and global feature distribution.
- A novel global feature-level loss function is designed in the basis of H. Barlow's redundancy-reduction principle, enhancing the similarity between the compared features while minimizing the redundancy between the remaining components of these vectors, accommodating the utilization of two different architecture in the S-T KD scheme.

The rest of the paper is organized as follows. Section 2 presents an overview of the recent deep learning methods related to OffSV problem. Section 3 describes thoroughly the proposed FKD method through the Student-Teacher architecture. Section 4 presents the experimental results investigating many different KD schemes and finally Section 5 provides discussion and conclusions.

## 2 Related Work

Training a feature extraction model for Offline Signature Verification (OffSV) is typically a separate learning task from the main verification task, as the identity and data of users involved in the operational phase are not available during the model's training phase. This provides flexibility in designing an efficient and practical OffSV system, resulting in a multitude of developed methods. Deep learning schemes have demonstrated effective performance in OffSV, mainly as feature extractors (Hameed et al., 2021). The main points of any deep learning based OffSV system could be summarized on the CNN architecture, the design strategy, and the multi-task learning mechanism. To achieve feature learning, a variety of CNN architectures are employed using either a WI or WD approach. The choice of architecture depends on both the user's requirements and the available signature data. Several OffSV datasets are available, as detailed in a recent survey by Diaz et al. (Diaz et al., 2019). In addition, a plethora of strategies have been developed to effectively capture the underlying signature information. The Siamese concept has a prominent position among these strategies since it is well-suited to the verification problem, having two inputs to compare two patterns and one output whose state value corresponds to the similarity between the two patterns (Bromley et al., 1994). Finally, the multi-task learning enjoys high popularity in the OffSV field due to its easy and effective implementation. The multi-task approach begins with a first task that acts as primary learning, while additional learning task(s) fine-tuned specific characteristics on the feature representations of signatures (Hafemann et al., 2017b). A taxonomy of deep OffSV methods according to the involved CNN architecture is attempted on Table 1, including information about the respective strategies and presence of multi-task stages.

The SigNet architecture, based on the AlexNet CNN topology, is dominant in the OffSV field (Krizhevsky et al., 2012). It was initially designed for writer identification, with the aim of distinguishing between signatures of different writers using only genuine signatures (Hafemann et al., 2017a). However, the architecture has since undergone various modifications, resulting in several versions of SigNet that differ mainly in the dimensions of the extracted features, such as the so-called thin SigNet (Hafemann et al., 2017a), R-SigNet (Avola et al., 2021), and SigNet-SPP (Hafemann et al., 2018). Also, different multi-loss settings have been employed as objective functions during training of SigNet. In these settings, the primary loss is responsible for associating signatures with their respective users, while additional loss terms are used either for detecting forgeries, resulting in the SigNet-F version (Avola et al., 2021; Hafemann et al., 2017a), or combined with other metric learning functions to form the Multi-Loss Snapshot Ensemble (MLSE) method (Masoudnia et al., 2019). To leverage the benefits of the SigNet-F feature extractor, post feature management methods are applied, such as the Dichotomy Transformation in the dissimilarity space (Souza et al., 2020) and the feature augmentation techniques to enhance the performance of the classifiers (Maruyama et al., 2021). The Siamese scheme is also formulated using the SigNet's architecture in its identical subnetworks (Dey et al., 2017). Building upon this, SigNet is utilized in multi-stage frameworks, either when it is initially trained to distinguish between signatures of different writers and subsequently re-trained using the contrastive loss function (Viana et al., 2023) or when it is initially trained with handwritten text data and then is used as the baseline model for training an additional contrastive loss layer at the top of the net (Tsourounis et al., 2022). The contrastive loss is the most common similarity ranking function in Siamese schemes and its objective is to learn such an embedding space in which similar sample pairs are pulled together while dissimilar ones are pushed apart (Hadsell et al., 2006). An extension of the Siamese concept is the triplet loss, which is composed of: an anchor, a positive sample from the same class, and a negative sample from a different class. In this case, the goal is to minimize the distance between the anchor and positive sample while maximizing the distance between the anchor and negative sample in the embedding space (Schroff et al., 2015; Weinberger & Saul, 2009). Beyond that, the dual triplets (or quadruplet) can also be used, which include two negative samples in addition to the anchor and positive samples (Chen et al., 2017). In the case of OffSV using SigNet model, the first negative sample is a random forgery and the second negative sample is a skilled forgery signature (Wan & Zou, 2021).

Many CNN architectures for OffSV systems are utilized as identical streams of joint embedding (i.e., Siamese) frameworks that rely on contrastive representation learning. One such architecture is the DenseNet (Huang et al., 2017) including squeeze-and-excitation blocks (SE) (Hu et al., 2018) with (Liu et al., 2021) and without (Liu et al., 2018) spatial pyramid pooling (SPP) for the calculation of the global feature. The SPP layer is also compiled with the custom CNN architecture, named Position-Dependent Siamese Network (PDSN), to model the local similarity between signatures (Lai & Jin, 2018), while a custom CNN equipped with an inception layer (Szegedy et al., 2015), named Siamese Convolutional Inception Neural Network (SCINN), is utilized to capture signature details (Ruiz et al., 2020). In the work of Parcham et al. (Parcham et al., 2021), the Capsule Neural Network (CapsNet) produces the final feature embeddings of the overall Siamese scheme and the resulting composite backbone architecture with the hybrid CNN-CapsNet models is named CBCapsNet. In this study, a plethora of CNN architectures are evaluated along with the CapsNet and thus, networks from the families of VGG (Simonyan & Zisserman, 2014), In/Xception (Chollet, 2017; Szegedy et al., 2015), ResNet (He et al., 2015; Szegedy et al., 2017; Xie et al., 2017), DenseNet (Huang et al., 2017), MobileNet (Howard et al., 2017), and NASNet (Zoph et al., 2018) were investigated under the proposed Siamese set up. The triplet-based learning is also performed in the OffSV problem using the VGG-16 model (Rantzsch et al., 2016) as well as the ResNet-18 and DenseNet-121 models (Maergner et al., 2019). In more complex configurations with custom CNN architectures for OffSV, the use of signature pairs can take many forms. For example, the work of Lu et al. (Lu et al., 2021) proposed the use of a smooth double-margin loss as an inventive extension of the contrastive loss while the work of Zhu et al. (Zhu et al., 2020) proposed a CNN equipped with fractional max pooling function as long as the contrastive and triplet losses are formulated with the novel point-to-set (P2S) similarity metric. The Deep Multitask Metric Learning (DMML), created by Soleimani et al. (Soleimani et al., 2016) as a multi-task learning version of Discriminative Deep Metric Learning (DDML), has a shared layer for all the writers that is followed by separated layers which belong to each writer independently and the overall topology is optimized using the relevant signature pairs. In addition, both the Inverse Discriminative Network (IDN) and the Multiple Siamese Net (MSN) utilize the original image (i.e., with white background and gray signature strokes) and the inverse version (i.e., with black background and gray signature strokes) of each signature of the input pair and through pairwise connection of its four different streams providing three (Wei et al., 2019) or four (Xiong & Cheng, 2021) verification scores that combined for the final decision. Finally, in an altered direction, a pair of grayscale signatures is fed into a custom CNN architecture as a two-channel input image to incorporate the similarity between the two signatures implicitly in the encoding process (Yilmaz & Öztürk, 2018).

Before their final use as feature extractors for signature images, popular CNN architectures are often trained following a different strategy, specifically in writer identification tasks, rather than the Siamese concept. The multi-task approach is commonly adopted in many cases, either as a multi-stage process where pretraining serves as coarse initialization for the network before the main training process specific to the method, or as a multi-loss implementation where multiple loss functions are optimized together to balance multiple objectives. For the OffSV problem, the ResNet-8 is pretrained with auxiliary Persian handwritten text images in the writer identification task and next either is used as a fixed feature extractor or is fine-tuned with signature images of the target domain (Mersa et al., 2019). Following a similar rationale, the CNNs are initially pretrained on the general imagenet dataset with millions of training images and then fine-tuning is performed on a single signature dataset under the writer identification problem to harness the effectiveness of the extracted vectors from the models such as VGG-16 (Engin et al., 2020), ResNet-50 (Engin et al., 2020; Younesian et al., 2019) and GoogLeNet (Jain et al., 2021). Likewise, the pretrained (on imagenet dataset) models of VGG-16, VGG-19, ResNet-50, and DenseNet-121 feed with feature representations the CapsuleNet that operates as the final verification classifier and the whole system is trained in an end-to-end manner using signatures (Yapıcı et al., 2021). Also, the custom topology combining two streams of convolutional processes in the work of Zheng et al. (Zheng et al., 2021) is pretrained on the signature writer identification task and subsequently the CNN is trained to capture micro deformations. Unlike the previous multi-stage approaches, the Shariatmadari et al. (Shariatmadari et al., 2019) trained their deep architecture utilizing a multi-loss approach combining two losses emerged from three CNN streams in different sizes of convolutional layers, while the proposed approach is based on Hierarchical One-Class CNN (HOCCNN) that trained only with genuine signatures from different feature levels. Furthermore, the CNN structure named Large-Scale Signature Network (LS2Net) with the class-center based classifier addresses the writer identification problem using the class centers -by averaging the extracted features of each class- and the 1-Nearest Neighbor classifier (Çalik et al., 2019).

The Recurrent Neural Networks (RNNs) are a type of neural network that are well-suited for processing sequential data. In the context of signature verification, RNNs can be used to analyze signature images by dividing them into segments and treating each segment as a separate time step in a sequence. There are several ways that RNNs can be applied to this task. One approach is to simply design geometrical windows on the pixel domain, as described in (Ghosh, 2020). Another approach is to use Local Binary Patterns (LBP) coded image windows, as described in (Yılmaz & Öztürk, 2020). These windows can be processed by a Bidirectional Long Short-Term Memory (BiLSTM) network, which is a type of RNN that is able to analyze the input data in both forward and backward directions. In a simpler implementation, a CNN can be used for feature extraction, with the output of the CNN being fed into a BiLSTM to classify the signature as genuine or forged (Longjam et al., 2023). The Static-Dynamic Interaction Network (SDINet) is another method for incorporating sequential information into static signature images by assuming pseudo dynamic processes in the static image (Li, Wei, et al., 2021b). It does this by uniformly dividing the feature maps of the signature into rows and columns, with each row or column representing a dynamic unit in the signing process. Thus, the static feature maps are converted into sequences based on the part-by-part nature of the signing process.

In the field of signature verification, there have been alternative proposed approaches that deviate from the usual line of research that was described above. One such approach is to use an autoencoder to generate forgery signatures from the genuine ones, where the encoder model is utilized to extract features from signature images (Prajapati et al., 2021). In the same vein, another approach is to use an Adversarial Variation Network (AVN), as proposed in the work of Li et al. (Li, Wei, et al., 2021a). The AVN exploits a variation consistency mechanism to train a discriminative model for signature authentication that is more robust than a typical Generative Adversarial Network (GAN). The AVN's feature extractor and discriminator are equipped with a variator that slightly perturbs the colors or intensities of the signature images to produce variants that should not affect the verification decision. Additionally, adversarial examples, which are intentionally designed to mislead a classifier, can pose a challenge for OffSV systems, as they can cause misclassification (Hafemann et al., 2019; Li, Li, et al., 2021). Finally, Graph Neural Networks (GNNs) have been applied to the problem of OffSV for the first time in the work of Roy et al. (Roy et al., 2021) and the transformer structure has been introduced as a feature extractor for signature images by Ren et al. (Ren et al., 2023). Both of these approaches show promising results.

Recently, the Self-Supervised Learning (SSL) approach has been introduced for the OffSV domain. Two SSL approaches have been developed for this task. The first approach involves pretraining a ResNet-18 model by minimizing the cross-correlation matrix between compared features. The resulting model is then used as a fixed feature extractor (Manna et al., 2022). The second approach involves pretraining an image reconstruction network with an encoder-decoder topology. The encoder (ResNet-18) is then finetuned using a dual triplet loss, and the resulting model is used as a feature extractor (Chattopadhyay et al., 2022). Differently from these SSL approaches where supervisory signals are obtained from the data itself, we propose a KD method for the training of the OffSV feature extractor where the process is supervised from a teacher model. Hence, we leverage prior knowledge by having the student model use an existing efficient CNN model for signature encoding. Additionally, we utilize handwritten text data to transfer the knowledge from the teacher to the student and not signatures, contrary to the aforementioned SSL works that rely on signature samples from the same datasets for achieving descent performance. Although both methods, SSL and FKD, utilize loss functions based on the cross-correlation matrix of global features, in the proposed scheme the FKD loss function is tailored to the KD concept instead of feature similarity.

*Table 1: A taxonomy of recent deep learning-based OffSV systems.*

| CNN involved Architecture | | Strategy | Multi-task | References |
|---|---|---|---|---|
| **SigNet (AlexNet)** | SigNet, thin SigNet | Identification | - | (Hafemann et al., 2017a) |
| | SigNet-F | 2-term Loss | Multi-Loss | (Hafemann et al., 2017a) |
| | SigNet | Siamese | - | (Dey et al., 2017) |
| | SigNet-SPP, fine-tuned | 2-term Loss & fine-tuning | Multi-Loss | (Hafemann et al., 2018) |
| | SigNet (MLSE) | 3-term Loss | Multi-Loss | (Masoudnia et al., 2019) |
| | SigNet-F | Dichotomy Transformation | Multi-stage | (Souza et al., 2020) |
| | SigNet-F | Feature Augmentation | Multi-stage | (Maruyama et al., 2021) |
| | R-Signet-F | 2-term Loss | Multi-Loss | (Avola et al., 2021) |
| | SigCNN | Dual Triplets | - | (Wan & Zou, 2021) |
| | SigNet-CoLL | Contrastive Layer (CoLL) | Multi-stage | (Tsourounis et al., 2022) |
| | SigNet | Multi-task Contrastive Learning | Multi-stage | (Viana et al., 2022, 2023) |
| **ResNet** | ResNet-8 | Auxiliary data | Multi-stage | (Mersa et al., 2019) |
| | ResNet-18 | Pretraining Identification + Triplets | Multi-stage | (Maergner et al., 2019) |
| | ResNet-50 | Pretraining Imagenet + Active Learning | Multi-stage | (Younesian et al., 2019) |
| | ResNet-50 | Pretraining Imagenet + Identification | Multi-stage | (Engin et al., 2020) |
| | ResNet-18 | SWIS: Self-Supervised Pretraining + Contrastive | Multi-stage | (Manna et al., 2022) |
| | ResNet-18 | SURDS: Self-Supervised Pretraining + Dual Triplets | Multi-stage | (Chattopadhyay et al., 2022) |
| **VGG** | VGG-16 (reduced) | Pretraining Identification + Triplets | Multi-stage | (Rantzsch et al., 2016) |
| | VGG-16 | Pretraining Imagenet + Identification | Multi-stage | (Engin et al., 2020) |
| **DenseNet** | DenseNet-36 | Multi-region + Siamese | - | (Liu et al., 2018) |
| | DenseNet-121 (MCS) | Pretraining Identification + Triplets | Multi-stage | (Maergner et al., 2019) |
| | Mutual Signature DenseNet-36 (MSDN) | Multi-region, SPP + Siamese | - | (Liu et al., 2021) |
| **InceptionNet** | Convolutional Inception NN (SCINN) | Signature Synthesis + Siamese | - | (Ruiz et al., 2020) |
| | GoogLeNet | Pretraining Imagenet + Identification | Multi-loss/ -stage | (Jain et al., 2021) |
| **CapsuleNet** | VGG-16/19, DenseNet-121, ResNet-50 + CapsNet | Signature Augmentation + Pretraining Imagenet + end-to-end Verification | Multi-stage | (Yapıcı et al., 2021) |
| | VGG-16/19, ResNet-50/101/152, In/Xception, InceptionResNet, MobileNet, NASNet + CBCapsNet | Siamese | - | (Parcham et al., 2021) |
| **Custom CNN** | Shared layers followed by separated layers (DMML) | DDML with User-specific layer + Pairs | Multi-stage | (Soleimani et al., 2016) |
| | 2-channel CNN | 2-channel input Pair | - | (Yilmaz & Öztürk, 2018) |
| | CNN + PSDN | Siamese | Multi-Loss | (Lai & Jin, 2018) |
| | 4-stream CNN | Multi-Path + Pairs (IDN) / (MSN) | - | (Wei et al., 2019; Xiong & Cheng, 2021) |
| | HOCCNN | Hierarchical one-class Learning | Multi-Loss | (Shariatmadari et al., 2019) |
| | LS2Net | 1-Nearest Neighbor (1-NN) classification task by using the class-centers | - | (Çalik et al., 2019) |
| | CNN with fraction max pooling | Point-to-Set (P2S) Similarity | - | (Zhu et al., 2020) |
| | 2-stream combined CNN | Pretraining Identification + micro-Deformations Learning | Multi-stage | (Zheng et al., 2021) |
| | cut-and-compare Net | Segmentation, Comparison + Pairs | - | (Lu et al., 2021) |
| | SDINet | conversion of static feature maps into sequences | - | (Li, Wei, et al., 2021b) |
| **RNN** | LSTM/ BiLSTM | Spatial segments + Identification | - | (Ghosh, 2020) |
| | Recurrent Binary Pattern – BiLSTM | LBP coded windows + Identification | - | (Yılmaz & Öztürk, 2020) |
| | CNN with BiLSTM | Hybrid CNN-BiLSTM verification | - | (Longjam et al., 2023) |
| **Autoencoder** | custom 6-layer CNN | Utilization of encoder model | - | (Prajapati et al., 2021) |
| **AVN** | VGG (inspired from) | Variation consistency mechanism | - | (Li, Wei, et al., 2021a) |
| **GraphNN** | GLCM-GNN | Node Classification | - | (Roy et al., 2021) |
| **Transformer** | two-channel and two-stream (2C2S) transformer | squeeze-and-excitation (SE) operation between two standard Swin Transformer blocks + Pairs | - | (Ren et al., 2023) |
| **Adversarial attack** | adversarial examples | adversarial characterization / adversarial perturbations | - | (Hafemann et al., 2019; Li, Li, et al., 2021) |

# 3 Proposed Method

## 3.1 Harnessing Knowledge through Distillation

The efficiency of CNNs in the modern Deep Learning era is founded on large and annotated training datasets and thus, the amount and quality of both data and labels is mission-critical. The most popular approach for reducing the amount of labeled training data without affecting the performance too much, is by employing prior knowledge from a source domain with an abundance of training data on a similar task. Then, transfer knowledge is performed from the source task to enable the learning on the target task utilizing the same network sequentially (Hussain et al., 2019). Towards a similar goal but with on a slightly different line, Knowledge Distillation (KD) is a method for transferring information from one network to another network whilst training constructively (Ba & Caruana, 2014; Hinton et al., 2015). The most prominent setting of KD is a Student-Teacher (S-T) scheme, where the knowledge is transferring from a "Teacher (T)" model to a "Student (S)" model and in this manner, the teacher CNN is supervising the training of the student CNN. Since the knowledge from the teacher reflects a more general type of information that could be expressed through many representations, there is no commonly agreed theory as to how knowledge is transferred. Therefore, various forms of KD methods are developed covering different aspects, like the types of distillation, the quality measures of knowledge, the design of S-T architecture, etc. A detailed survey on KD and S-T learning methods can be found in (Gou et al., 2021; Wang & Yoon, 2022).

Knowledge often refers to the learned weights and biases, although there is a diversity in the sources of knowledge in a CNN. Typically, the two principal sources of knowledge in a CNN model are, the output prediction score, known as logits, and the activations of intermediate layers, known as hints. Since the soft logits represent the class probability distribution, the knowledge from teacher's model is shifted to the student's model by learning the class distribution via softened softmax (also called "soft labels"), where each soft label's contribution is controlled using a parameter defined as temperature (Ba & Caruana, 2014; Hinton et al., 2015). The main idea is that the student model will learn to mimic the responses of the teacher model and not only the hard class predictions. However, since CNNs are compositional models that organize the information hierarchically, they could learn multiple levels of feature representation with increasing abstraction (Bengio et al., 2014) and thus, the knowledge derived from the intermediate layers of a teacher model could provide favorable information. Like so, the goal of this type of KD (Feature-based KD), is matching the internal representations between student and teacher models. Supplementary to the above sources, the knowledge that captures the relationship between different activations and neurons -from one or more locations of features along the network- can also be used to train a student model. Next, we present a detailed description of the used FKD, explaining how knowledge is measured and how the information is transferred from teacher to student through the proposed loss functions.

## 3.1.1 Geometric Regularization through Local Activation Features

The local features are the activations from intermediate layers of a CNN, meaning that they are the output of a hidden layer that constitute a Feature Map (FM). The FM is an intermediate representation generated from a convolution layer and thus, includes local information since each entry of FM highlights only a local neighborhood of input pixels. In an S-T FKD scheme, the teacher's intermediate representations supervise the training of the student model, so to learn feature representations that match some qualities of the respective teacher's predictions.

Given its spatial structure, a FM can be considered as a set of multidimensional vectors representing local features depth wise. The overall affinity between two sets of multidimensional data (feature vectors) can be measured through a similarity or dissimilarity function, formulated from either statistical or geometrical perspective. According to the statistical approach, the distance between two sets of feature vectors is related to the dissimilarity between the underlying distributions from which the vectors are derived. On the other hand, the geometrical approach assumes that the data from each set of vectors are lying on a low-dimensional manifold inside the feature space and thus, the distance can be defined as a measure of the dissimilarity between geometrical properties of the corresponding manifold structures. In (Theodorakopoulos et al., 2016), a manifold-to-manifold distance is introduced based on the notion of reordering efficiency of the neighborhood graphs representing the manifolds of local features. Following, in (Theodorakopoulos et al., 2020, 2021) this distance was extended to an efficient Feature-based Knowledge Distillation (FKD) technique through a geometric regularization of local activations within an S-T framework. Consequently, the local manifold-based regularization incentivizes a student CNN to create local features that resemble, in overall geometry, to those of a teacher model at several layers with matching spatial resolutions. In this work, we employ a FKD approach which is based on the above-mentioned manifold-to-manifold distance, regularizing the activations in several intermediate layers of student model via the respective activations in the teacher network.

### *Geometric Distillation*

Let us consider a Feature Map (FM) of size $H \times W \times D$, where *H* and *W* correspond to its spatial size (Height and Width) while the Depth size *D* denotes the number of channels. It consists of *N=H*W* feature vectors, each one having dimensionality of *D* (i.e., $x_j \in \mathbb{R}^{1 \times D}$ is a channel-wise feature vector with D elements, one for each pixel location j=1,…,N). Thus, a FM is a set of N feature vectors in a feature space of size D. Hence, the dissimilarity between two feature maps extracted from two CNNs could be measured via a manifold-to-manifold distance metric between the local activation manifolds, at corresponding layers of the two models with one-to-one correspondence between samples of the two compared sets. This happens in the case where the two compared feature maps have the same spatial size (H&W), independent of the feature dimensionality (i.e., the value of D). The neighboring relations within each feature set can be encoded by a Minimal Spanning Tree (MST), which in the form of a minimalistic backbone, connecting the nodes representing feature vectors (Kruskal, 1956). In such case, neighborhoods can be defined via a geodesic radius around each node on the MST. The MST was used in such setting because it is less prone to topological short-circuits and thus, generating neighborhoods whose affinities are more indicative of the underlying manifolds' features (Zemel & Carreira-Perpiñán, 2004). Finally, a measurable quantity of local affinity for each FM's vector can be obtained with the Neighborhood Affinity Contrast (NAC) (Theodorakopoulos & Tsourounis, 2023). The NAC measures the ratio of the sum of square Euclidean distances of a sample to all its neighbors, to the sum of distances to all the other samples of the set. Thus, NAC is an atypical measure of compactness of the local neighborhood of each sample. The, the NAC ratio (i.e., intra distance to inter distance) is calculated using the following formula:

$$NAC_M^{FM} = \frac{\sum_{j=1}^{N} dist_{ij} \cdot m_{ij}}{\sum_{j=1}^{N} dist_{ij}} \in \mathbb{R}^{1 \times N} \quad (1)$$

where the total number of FM vectors is $N = H * W$, the pairwise vectors' normalized square Euclidean distance is $dist_{ij} = \frac{\|x_i - x_j\|_2^2}{\sum_{j=1}^{N} \|x_i - x_j\|_2^2}$, $x_i \in \mathbb{R}^{1 \times D}$ and the neighborhood mask $M \in \{0,1\}^{N \times N}$ is based on the geodesic distance between the i-th and j-th nodes (i.e., feature vectors) on the MST with

$$m_{ij} = \begin{cases} 1, & Distance_{geodesic}^{MST}(i,j) \leq r \\ 0, & Distance_{geodesic}^{MST}(i,j) > r \end{cases} \quad (2)$$

with *r* a geodesic radius indicating the number of hops that define the neighbors of each node on the MST.

In this work, the neighborhood mask $M \in \{0,1\}^{N \times N}$ is computed only on the teacher's side ($Mt$) and once for each datum, as proposed in (Theodorakopoulos et al., 2020, 2021; Theodorakopoulos & Tsourounis, 2023), to force the student model's activations to mimic the neighboring relations, as expressed in the corresponding activations in the teacher's model. Therefore, the student's model is guided to produce local activation features with similar geometrical characteristics to those of the teacher model. The comparison between a feature map from the teacher CNN (FMt) and a feature map from the student CNN (FMs) -with equal spatial resolutions- is provided by the mean squared distance between the respective NAC vectors from the teacher's and student's FM, using the same neighborhood mask $Mt$. Therefore, the Geometrical Loss of the local features from the intermediate layers in the utilized S-T FKD scheme is defined as:

$$GeomL = \left\| NAC_{Mt}^{FMt} - NAC_{Mt}^{FMs} \right\|_2^2 \quad (3)$$

By designing in this way, multiple supervision connections between layers of the teacher and the student can be simultaneously implemented, adding the respective regularization terms in the overall loss function. Also, the geometrical regularization of local activations could work synergistically with any other KD loss as well as task-dependent loss terms. In our implementation, we utilize two connections between intermediate layers on the S-T scheme for geometric regularization of local activation features. Thus, there are two geometrical regularization terms targeting different layers of the networks, one in an early layer and one after the middle of the teacher's topology, connecting with layers equivalent in terms of spatial size to the student's topology (Figure 1).

## 3.1.2 Response Regularization through Global Features

Typically, when a CNN model is utilized for feature extraction, the extracted feature is provided by the penultimate layer, just before the classification layer. Since this feature is the network's response, optimized so to facilitate a classification task, it reflects the discriminative qualities learned by the model during training. Hence, if the global feature of a CNN model exhibits preferable characteristics, it is reasonable to try to teach the student model to imitate this directly. Fortunately, in a S-T scheme, the teacher's knowledge could be transferred to the student through the feature information of the respective responses, regardless of the classification task solved by the teacher and the respective data.

Assuming that the global features from the two models (teacher and student) have the same dimensionality, it is easy to formulate a function to compare them. In this work, the FKD through global features is incurred by minimizing the difference between the teacher response and student response using either the cross-entropy loss of the two temperature-scaled features, or a loss utilizing the cross-correlation matrix between the two feature vectors.

### *Response Distillation based on cross-entropy*

The softmax function transforms the input vector into a probability distribution promoting the highest value against others. Accordingly, the output values are restricted to the range of [0,1] with their sum being equal to 1, whilst the larger values are intensified, and the lower values are denoted. One effective calibration technique for rescaling the output values to increase the sensitivity of low probability candidates is the temperature scaling (Ba & Caruana, 2014; Hinton et al., 2015). The softmax with temperature parameter softens the distribution by penalizing the larger logits more than the smaller logits and thus, more probability mass will be assigned to the smaller logits. This characteristic could be very beneficial for our case, where high-dimensional feature vectors (and not class predictions) are utilized directly in the loss function. The softmax with temperature parameter $\tau$ for an input vector $f \in \mathbb{R}^{1 \times K}$, where $K$ is the feature dimensionality, is calculated as:

$$p_i = \frac{e^{(f_i/\tau)}}{\sum_l^K e^{(f_l/\tau)}} \qquad (4)$$

The imitation of teacher's output by the student model is driven by the minimization of cross-entropy between the temperature scaled features extracted from two CNNs respectively. Thus, the Response Loss of the final features with the temperature-scaled cross-entropy function (T-CE) in the S-T FKD scheme is defined as:

$$RespL_{\text{TCE}} = -\sum_i^K ( q_i^t \cdot \log(p_i^s) ) \qquad (5)$$

where $p_i^s$ and $q_i^t$ are the final extracted features after temperature softmax from the student and the teacher respectively while K equals to the features' dimensionality.

In our implementation of S-T FKD framework, the above Response regularization term is evaluated either in conjunction with classification loss or other KD losses, or as a single loss term of the training procedure.

## Response Distillation based on cross-correlation

The cross-correlation between two different signals can be used as a technique for comparing two signals. A commonly used extension of the simple cross-correlation is normalized cross-correlation which can detect the correlation of two signals with different amplitudes. The cross-correlation matrix of two vectors in the $\mathbb{R}^K$ space is a matrix with elements the cross-correlations of all pairs of elements of the vectors. In particular, the cross-correlation matrix $C$ between the normalized responses from the teacher ($z^t \in \mathbb{R}^{1 \times K}$) and the student ($z^s \in \mathbb{R}^{1 \times K}$) is computed by the following formula:

$$C_{ij} \triangleq \frac{\sum^K z_i^s \cdot z_j^t}{\sqrt{\sum^K (z_i^s)^2} \sqrt{\sum^K (z_j^t)^2}} \quad (6)$$

where $C_{ij}$ is the correlation between i-th element ($z_i^s$) of normalized student's vector ($z^s \in \mathbb{R}^{1 \times K}$) with the j-th element ($z_j^t$) of normalized teacher's vector ($z^t \in \mathbb{R}^{1 \times K}$), while the above relation could be also applied to vectors normalized via z-score standardization across batch, during the training of the proposed S-T framework.

An objective function which tries to make two feature vectors similar while reducing the redundancy between their components, can be expressed by enforcing their cross-correlation matrix as close to the identity matrix as possible. For this purpose, the Barlow Twins loss function (Zbontar et al., 2021) has been proposed as a self-supervised learning approach, comparing the embeddings of two distorted versions of an input image by the same network. In our work, we exploit a similar logic, but comparing the features extracted from two different networks instead, (i.e., teacher and student) for the same input image. Thus, the cross-correlation matrix between the feature responses of the teacher and student CNNs is computed, creating an S-T FKD scheme where the Response Loss for the final features is defined as:

$$RespL_{\text{BT}} = \sum_i^K (1 - C_{ii})^2 + \lambda \sum_i^K \sum_{j \neq i} (C_{ij})^2 \quad (7)$$

with a trade-off parameter $\lambda \geq 0$ that controls the importance between the two terms of the loss, with the first term trying to pull the diagonal elements of the cross-correlation matrix towards 1 and the second term trying to minimize the off-diagonal elements. The features are centered and normalized to unit variance along the batch dimension before the calculation of the cross-correlation matrix.

In addition to the aforementioned loss term, we also evaluate a more relaxed version that offers some additional degrees of freedom to the student model's response, and is more compatible with the context of S-T distillation. Whilst the Response loss in Barlow Twins tries to match the response in the i-th element of student's vector to the i-th element of teacher's vector, the relaxed version tries to match the i-th element of student's vector with the element of teacher's vector with which it has the maximum correlation. The rationale behind this option is that since the different architectures of the student and teacher networks do not share parameters as in a Barlow Twins self-supervision setting, the two models could produce similar responses but within an arbitrary permutation of their features' elements. Hence, the utilized loss should facilitate such behavior and not necessarily enforce element-wise correspondence between the student's and teacher responses. Therefore, we opted for a loss function that accentuates the maximum correlation for each feature component. Ultimately, the proposed Response Loss of the final features in the S-T FKD scheme is defined as:

$$RespL_{\text{BC}} = \sum_i^K \left(1 - \max_j(C_{ij})\right)^2 + \lambda \sum_i^K \sum_{j \neq argmax_j(C_{ij})} (C_{ij})^2 \qquad (8)$$

As in the previous case, the parameter $\lambda \geq 0$ is a coefficient that adjusts the balance between the invariance term which enhances the maximum activations between the compared features, and the redundancy reduction term that decorrelates all the remaining components of the features. For this relaxation of the Barlow loss function we will use the term Barlow Colleagues (BC), in contrast to the unmodified Barlow Twins (BT).

## 3.2 Building the Student-Teacher Knowledge Distillation (S-T FKD) Architecture

Figure 1 presents the proposed S-T FKD scheme which distils the knowledge via an offline approach, where the teacher model is fixed, and the knowledge is transferred using both the feature maps of intermediate layers and the final features, to leverage the local and global information respectively.

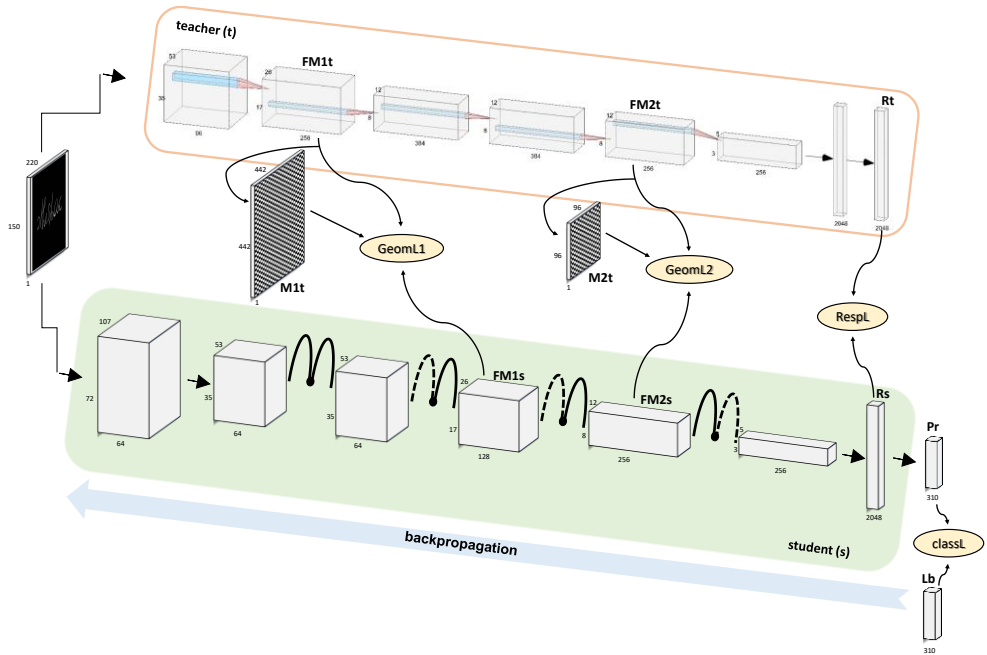

*Figure 1: The S-T scheme presents the produced representations as data pass the sequential operations of student and teacher CNNs. The student CNN has a fully connected layer at the end of its topology that serves the classification task (classL) using the predicted classification scores (Pr). The teacher model provides the neighborhood masks (M1t and M2t) for any given Feature Map (FM1t and FM2t) as well as the final feature vector response with 2048 elements (Rt). Two Feature Maps (FM1s and FM2s) as well as the final extracted feature (Rs) of the student model participate in the FKD. The FKD is implemented utilizing the geometrical regularization terms from intermediate layers (GeomL1 and GeomL2) and the response regularization term from the final extracted features (RespL). Thus, the overall multi-loss function combines the FKD regularizations together with the classification loss to supervise the training of the student CNN.*

*Table 2: Architectures and activations' size for teacher (SigNet) and student (ResNet18) models.*

| SigNet | | ResNet-18 | |
| --- | --- | --- | --- |
| Layer/Block Name | Activation size | Layer/Block Name | Activation size |
| Input | 150×220×1 | Input | 150×220×1 |
| conv1 | 35×53×96 | conv | 72×107×64 |
| pool1 | 17×26×256 | pool | 35×53×64 |
| conv2 (FM1t) | 17×26×256 | block1 | 35×53×64 |
| pool2 | 8×12×256 | block2 (FM1s) | 17×26×128 |
| conv3 | 8×12×384 | block3 (FM2s) | 8×12×256 |
| conv4 | 8×12×384 | block4 | 3×5×256 |
| conv5 (FM2t) | 8×12×256 | fc1 (Rs) | 2048 |
| pool5 | 3×5×256 | fc2 | #classes |
| fc6 | 2048 | | |
| fc7 (Rt) | 2048 | | |
| fc8 | #classes | | |

First, to form the S-T architecture, we utilize as the teacher CNN the original SigNet feature extractor proposed by Hafemann et al. in (Hafemann et al., 2017a). The used SigNet model provides a feature representation for any input image and not a predicted classification result whereas a classification score depends on the number of classes of the corresponding training dataset. Thus, for any input image, the SigNet produces a feature vector of 2048-dimensions. While the teacher SigNet is built on the base of AlexNet architecture, the student CNN follows a more modern architecture based on the ResNet-18 topology. Taking advantage from its residual skip connections via the four residual blocks, the student model is much deeper than the teacher model even though they have approximately equal number of learned parameters given the addendum of one fully connected layer with 2048 neurons in the student CNN as its penultimate layer for feature extraction before the final classification layer. The inclusion of the fully connected layer for feature extraction has as input the feature map of the previous layer without using any spatial pyramid layer. The architectures and sizes of activations both for the teacher and student models are presented in Table 2.

Secondly, the geometrical regularization requires to define the positions of distillation in the two models while the unique meeting condition is the same spatial resolution of the two connected feature maps. We select to utilize the final representation for each spatial resolution assuming that incorporates its optimal knowledge. For an input image of 150 x 220 pixels, the spatial pixel resolutions during the pass inside the teacher SigNet model are 35 x 53 (conv1-bn1-relu1), 17 x 26 (maxpool1 or conv2-bn2-relu2), 8 x 12 (maxpool2 or conv3-bn3-relu3, conv4-bn4-relu4, conv5-bn5-relu5), and 3 x 5 (maxpool5). Given the four different spatial resolutions founded along teacher SigNet's layers, the output from the first convolutional layer (i.e., 35 x 53) presumably captures primordial information and the smallest representation (i.e., 3 x 5) is degenerated; thus both they are deregistered. Hence, two spatial sizes, the 17 x 26 and 8 x 12 resolutions, are remaining. We opt to make use of the output after the consecutive operations of convolution, batch-normalization, and relu non-linearity supposing that this representation includes the best possible information as well as makes easier the correspondence with the output of a residual block in the student model. In this manner, the spatial resolutions lead to utilize the feature map representations produced as the output of the second and third residual block in the student's ResNet-18 model. Finally, the one geometrical regularization (*GeomL1*) is utilized the teacher's FM of size 17 x 26 x 256 and the student's FM of size 17 x 26 x 128 while the other geometrical regularization (*GeomL2*) uses the volumes with size 8 x 12 x 256 from the teacher and the student respectively. The response regularization (*respL*) is computed with the final extracted features, which are provided from the CNNs' layer when the classification output score is removed (i.e., if the classification task is not considered) and incorporate the total global knowledge of the network. Thus, the feature vector from the output of the feature extractor teacher (layer fc7 at SigNet model) and the feature vector from the fully connected layer (layer fc1 at ResNet-18 model) of the student CNN are utilized. Both features have 2048 elements and produced after the consecutive operations of fully convolution, batch-normalization, and relu non-linearity in the two models.

The Geometrical Loss (GeomL) of the intermediate feature maps (FM) from two different distillation positions (FM1t, FM1s and FM2t, FM2s) and the Response Loss (RespL) of the final extracted features (Rt, Rs) are implemented as regularization terms of the S-T FKD training. These are considered together with the typical (cross-entropy) Classification Loss (classL or CL) of the ground truth labels (Lb) and predicted classes (Pr), to form an overall multi-loss function that supervises the training of the student CNN. Thus, the overall multi-loss function of the S-T scheme is defined as:

$$\mathcal{L} = l_1 \cdot GeomL1 + l_2 \cdot GeomL2 + g \cdot RespL + c \cdot classL \qquad (9)$$

with coefficients $l_1, l_2, g, and\ c$ representing the weights for each term that contributes to the overall loss, where $c = 1$ to allow compiling relative relations with the other terms.

# 4 Experimental Evaluation

## 4.1 Datasets

### 4.1.1 Offline signature datasets for evaluation

The CNN models trained via the FKD processes are applied on three most popular OffSV datasets to assess their efficiency. As mentioned above, in OffSV setting, the CNN models are utilized for feature extraction, while WD classifiers undertake the final signature verification stage. The GPDS300GRAY, MCYT75, and CEDAR offline signature datasets are evaluated to measure the performance of the models as feature extractors. The three datasets include signatures scanned from documents as grayscale images.

The GPDS300GRAY is a standard subset of GPDS960 corpus (Digital Signal Processing Group) with 253 writers and each of them has 24 genuine and 30 forgeries signatures (Blumenstein et al., 2010; Ferrer et al., 2012; Vargas et al., 2007). The forgeries signatures marked as skilled since they made by 10 forgers from 10 randomly selected genuine specimens and the forger was allowed to practice the signature without time limit. Although the GPDS960 database is no longer publicly available due to the General Data Protection Regulation (EU) 2016/679 ("GDPR"), we utilize the GPDS300GRAY dataset only for evaluation to measure the performance and to accomplish comparisons with other works that report results on this dataset (Arab et al., 2023; Longjam et al., 2023; Viana et al., 2022).

The MCYT75 (Ministerio de Ciencia Y Tecnologia, Spanish Ministry of Science and Technology, MCYT-75 Offline Signature Baseline Corpus ("Database")) has 75 enrolled writers with 15 genuine and 15 forgeries signatures per writer (Fierrez-Aguilar et al., 2004; Ortega-Garcia et al., 2003). The forgeries generated by 3 different user-specific forgers and thus, they are skilled simulated signatures. The offline handwritten signature dataset MCYT75 is publicly available.

The CEDAR dataset (Centre of Excellence for Document Analysis and Recognition) includes 55 writers with 24 genuine and 24 forgeries signatures per writer (Kalera et al., 2004). The forgeries are a mixture of random, simple, and skilled simulated signatures since they are contributed by some writers of the dataset that asked to forge three other writers' signatures, eight times per subject. The offline CEDAR dataset is publicly available too.

## *Preprocessing*

The grayscale signature images are subjected to some simple preprocessing steps dedicated to normalization, noise removal and size correction, since scanned images may contain noise and also the methods require the images in a fixed size. The normalization process shares the same steps as many previous works on OffSV (Hafemann et al., 2017a, 2018; Viana et al., 2022, 2023), something that also enables fair performance comparisons. The preprocessing includes the following steps: gaussian filtering and OTSU thresholding to remove background noise, centering into a large blank canvas of a predefined size of 952 x 1360 pixels (Height x Width) -common for all datasets- by aligning the signatures' center of mass to the center of the canvas so as not to affect the width of strokes and to present the original aspect ratio, inverting the images to have black background and grayscale foreground by subtracting the maximum brightness (i.e., white value of 255), and resizing the images to the input size of the CNNs that is 150 x 220 pixels (Height x Width).

## 4.1.2 Handwritten Text data for Training in S-T configuration

Training in a S-T configuration is realized using text data that work as information carrier to transfer the knowledge from the teacher into the student. Taking into consideration the biometric qualities of handwriting, the handwritten text data from the auxiliary domain are processed by a specially designed procedure to create an auxiliary task. The auxiliary task was designed so as its data to resemble more to those of the target domain of handwritten signature images. The text data come from the publicly available CVL-database, where 310 writers fill in 5-10 lines of predefined text on page-forms (Kleber et al., 2013). The text data are processed according to the procedure proposed by Tsourounis et al. in (Tsourounis et al., 2022) to generate text images that resemble the distributions of signature images and use them as the training data of a CNN that solves a writer identification problem. In brief, the handwritten text documents are first converted to grayscale, then the lines of text are isolated as Solid Stripes of Text (SsoT), and finally, the SsoT are cropped into vertical intervals to generate the text images. Subsequently, the text images are preprocessed as signature images, following the same preprocessing steps detailed above. Given the findings in (Tsourounis et al., 2022) about the effects of cropping and canvas dimensions of the text data, the random choice of parameters is more favorable offering better generalization. Hence, in this work the utilized training text set is obtained by each SsoT with random aspect ratio and using all available canvas sizes. In the context of this work though, the increased computational load needed for each training image -given the estimation of the Neighborhood Affinity Contrast (NAC) for some Feature Maps (FMs) as well as the calculation of the MST from the side of the teacher- led us to reduce the training set by sampling one text image for each canvas. Ultimately, about sixty thousand training and twenty-five thousand validation images are used for the training of S-T schemes (the training set could be downloaded from the official repository of our work[2]).

## 4.2 Experimental Setup and Protocols

In the context of this work, the trained CNN models are utilized as feature extractors, and the produced descriptors of the input signatures are consumed by binary classifiers that distinguish the genuine from the forgery signatures. Hence, the generalization performance of the CNN models is measured using the evaluation metrics obtained from the classifiers in the verification task. In this work, we follow the Writer Dependent (WD) approach and thus, a Support Vector Machine (SVM) classifier is trained for each writer. The implementation of WD classifiers is based on the work of Hafemann et al. (Hafemann et al., 2017a). In this manner and for fair comparisons, we utilized the implementation provided in the official repository[3] of (Hafemann et al., 2017a) for the partition into training and test sets, the training of the classifiers, and the calculation of evaluation metrics.

### 4.2.1 Experimental Protocol

Initially, a number of genuine signatures for every writer, denoted as the number of reference signatures $N_{REF}$, is selected to form the training set for the SVMs. In this manner, for each writer's SVM, the positive training class consists of the reference signatures from the writer while the negative training class is composed of the reference signatures from all other writers of the evaluated dataset (also called random forgeries). The test set for each writer includes the remaining genuine signatures from the writer and a number of the corresponding skilled forgeries. Following the works of (Hafemann et al., 2017a; Maruyama et al., 2021; Viana et al., 2023), the number of training and testing samples used for evaluation on each dataset, is summarized in Table 3. Preferably, the number of genuine test samples would be equal to the test skilled forgeries (without diminishing the test set) for the two populations to have equal contributions to the error. The chosen number of the reference signatures per subject is in line with the most common experimental protocols in the literature (Hafemann et al., 2017b). For each experiment, ten (10) repetitions with WD classifiers trained using randomly selected splits of data are performed (with different reference signatures), and the results are presented in terms of the average and standard deviation values across these 10 iterations.

---

[2] https://github.com/dimTsourounis/FKD
[3] https://github.com/luizgh/sigver

*Table 3: The partition into training and test sets for the WD classifiers using the signature datasets.*

| OffSV Dataset | | | | Training set | | Test set | |
|---|---|---|---|---|---|---|---|
| Name | Writers | Genuine | Skilled Forgeries | Genuine (writer's $N_{REF}$) | Random Forgeries (other's $N_{REF}$) | Genuine (Rest) | Skilled Forgeries |
| GPDS300GRAY | 253 | 24 | 30 | 12 | 12 x 252 | 10 | 10 |
| MCYT75 | 75 | 15 | 15 | 10 | 10 x 74 | 5 | 15 |
| CEDAR | 55 | 24 | 24 | 12 | 12 x 54 | 10 | 10 |

## 4.2.2 Writer Dependent classifiers (WD SVM)

The WD classifiers were trained using soft margin binary Support Vector Machine (SVM) with Radial Basis Function (RBF) kernel, while the two associated hyper-parameters (cost parameter C and scaling parameter gamma), were set to constant values of $C = 1\ and\ \gamma = 2^{-11}$. Also, more weight to the positive class is used in order to correct for the class imbalance, given that the positive training class consists of only a few genuine signatures and the negative training class has much more signatures due to the usage of samples from many writers. So, the weight for the negative class is set to 1 and the weight for the positive class is the ratio of the number of negative training examples to the number of positive training examples.

## 4.2.3 Evaluation Metrics

The number of reference signatures specifies the genuine signatures of a writer, used to construct the positive class during the training of its corresponding SVM, while the negative training class is created from the reference signatures of all other writers of the dataset. After the training of an SVM, a decision threshold should be defined to distinguish any query test signature as genuine or forgery. Mainly one of two approaches is followed to determine the decision threshold; either utilizing all the available training signatures of the datasets (from all the users) or using just the training signatures that correspond to each specific user, in order to select the threshold closest to FPR = 1 – TPR (i.e., False Positive Rate equals to one minus True Positive Rate). The first approach sets one optimum global decision threshold (a posteriori) that is common for all the writers' SVMs, and the second approach sets user-specific thresholds by using the optimal decision threshold for each writer's SVM individually. For calculating the False Acceptance Rate (FAR: misclassifying a forgery as being genuine) and False Rejection Rate (FAR$_{skilled}$: misclassifying a genuine as being skilled forgery), the global decision threshold is used. The Equal Error Rate (EER) for each user is calculated considering only skilled forgeries (not random forgeries) when FRR equals to FAR$_{skilled}$ and using two forms, the global decision threshold to report EER$_{globalthreshold}$ and the user-specific threshold to report EER$_{userthreshold}$. In both cases, the reported EER is the average value from all the writers of the dataset and after the ten repetitions of experiment with different reference samples for every writer in each iteration. Finally, the mean Area Under the Curve (AUC) is often used as a metric measured on the ROC curves created for each user individually.

## 4.3 Implementation Details of S-T training

The only parameter of the utilized geometric regularization is the radius of *r* that define the neighborhood size on each respective MST. In the following, we use a radius of $r = 5$, a value resulted as the most reliable setting for good performance in a small set of preliminary experiments with the selected teacher model. Also, following the observation made in (Theodorakopoulos et al., 2020, 2021) that training is more stable when earlier layers have a smaller contribution in the overall loss than the deeper ones, we set the contribution coefficients $l_1 = 10 \ and \ l_2 = 100$ throughout the evaluation, since these values provided the good results in the same preliminary experiments. The response regularization is implemented in three different ways, using the cross-entropy loss of temperature scaled features (T-CE), the cross-correlation matrix of normalized features based on Barlow Twins loss (BT), and the novel version of the latter named Barlow Colleagues (BC). For the first case, we ran a search for the temperature factor $\tau$ as well as the coefficient of contribution $g$ and found the best results for $\tau = 10$ and $g = 0.001$. For the other two cases, the trade-off parameter $\lambda = 0.0001$ and coefficient $g = 0.0001$ are set after a grid search. Our observations suggest that a downscale (approximately two or three orders of magnitude) of the response distillation loss term relative to the classification loss, is beneficial to the overall performance. Additionally, the utilization of the regularization terms together with the classification term from the beginning of the training, produced better results than applying a warmup training with only the classification loss.

The S-T framework was trained using the Stochastic Gradient Descent (SGD) optimizer with initial learning rate of 0.01 which is reduced by a factor of by 10 every 20 epochs for a total of 60 epochs, using Nesterov Momentum with a momentum factor of 0.9. The batch size was 64 according to the maximum capacity of the utilized GeForce RTX 2070 GPU. Each S-T training took approximately 30 hours. Implementation of the proposed FKD method is available for download at the official repository[4].

## 4.4 Results and Analysis

### 4.4.1 Proof-of-concept (SigNet-to-SigNet)

A goal of this work is to demonstrate teacher-to-student knowledge transfer using exclusively auxiliary data. The first setting we investigated as a proof-of-concept is when the teacher and student models follow the same architecture. In this manner, we investigate only if the transfer of knowledge is effective, without any performance contributions stemming from architectural differences. In this experiment, the teacher model is a SigNet, which is trained on signature images (Hafemann et al., 2017a) and is not updated during S-T training, while the student model follows the SigNet's architecture with random parameters initialized using the Xavier sampling (Glorot & Bengio, 2010). The student model is trained on the task of text-based writer identification, utilizing different combinations of distillation losses along with the main classification loss. The evaluation of the trained student model was performed on the signature verification task, using WD classifiers that trained on the features extracted from layer fc7 of the student model and following the user threshold approach to calculate the performance metrics. Table 4 summarizes the obtained EER from the models trained in both standard identification task (CL), as on the various S-T training configurations. The performance of the teacher and the student models at their initial conditions is also reported as baseline performance.

---

[4] https://github.com/dimTsourounis/FKD

Table 4: Performance of the WD classifiers with user-threshold on the SigNet-to-SigNet FKD schemes.

| Method | Overall loss | | | | EER (user threshold) | | |
| --- | --- | --- | --- | --- | --- | --- | --- |
| | $l_1$ | $l_2$ | $g$ | $c$ | GPDS300GRAY $N_{REF}=12$ | MCYT75 $N_{REF}=10$ | CEDAR $N_{REF}=12$ |
| Teacher – SigNet (fixed) | - | - | - | - | 3.29±0.24 | 3.14±0.60 | 4.33±0.66 |
| Student – SigNet (Random W) | - | - | - | - | 9.53±0.39 | 10.71±1.17 | 11.95±0.81 |
| CL (w/o KD) | 0 | 0 | 0 | 1 | 4.28±0.17 | 8.27±0.67 | 3.91±0.60 |
| CL + KD: GEOM | 10 | 100 | 0 | 1 | 3.39±0.17 | 7.30±0.74 | 3.58±0.24 |
| CL + KD: T-CE | 0 | 0 | 0.001 | 1 | 4.11±0.19 | 8.26±1.42 | 2.94±0.34 |
| CL + KD: BT | 0 | 0 | 0.0001 | 1 | 4.46±0.38 | 4.76±0.97 | 2.94±0.34 |
| CL + KD: BC | 0 | 0 | 0.0001 | 1 | 3.78±0.24 | 7.92±0.81 | 3.78±0.56 |
| CL + KD: GEOM & T-CE | 10 | 100 | 0.001 | 1 | 3.37±0.19 | 7.49±1.20 | 3.02±0.34 |
| CL + KD: GEOM & BT | 10 | 100 | 0.0001 | 1 | 3.40±0.27 | 7.55±1.29 | 3.73±0.56 |
| CL + KD: GEOM & BC | 10 | 100 | 0.0001 | 1 | 3.17±0.16 | 7.02±1.26 | 3.51±0.32 |

As can be easily inferred by the results of Table 4, the utilization of any feature knowledge distillation (FKD) technique together with classification loss (CL) is beneficial in apposition to sole CL. Also, training only on the textual task alone (CL) produces a model with less discriminative features for the OffSV task, delivering inferior performance to the teacher model, but superior to randomly initialized model as expected. Since many experiments provide EER values with small differences, statistical tests for the ten repetitions in each setting using both Friedman and paired signed-rank Wilcoxon with $p<0.05$ were performed to clarify the comparisons against the teacher model. First, we can observe that the trained student model is statistically better in the CEDAR dataset for all the FKD schemes, while all variations are statistically on par with the teacher model in the GPDS dataset. Secondly, the results in the MCYT dataset are inconclusive both for the standard CL training and on the S-T schemes. An exception in the above is the case when classification loss and response distillation with BT loss (i.e., CL + KD: BT) is applied, where worse performance in GPDS dataset and better in MCYT are observed. An explanation behind this behavior could be that the architecture between teacher and student CNNs is the same. Finally, the large EERs in MCYT dataset meaning there is a trammel that degrade the performance, and this could be caused by the utilized text data that cannot adequately simulate the distribution of signatures on this dataset, with the limited capability of the student CNN having also a negative impact. In the following experiments, the same training regime is retained but the student CNN is changed from the AlexNet-based SigNet topology to the modern and efficient ResNet-18 architecture.

## 4.4.2 Model-to-Model Experiments (SigNet-to-ResNet)

Once the functionality of the proposed mechanism for knowledge transfer is established, our main goal is to train and evaluate new models with the ResNet-18 architecture. Additionally, we examine the effects of local and/or global distillation terms in conjunction with the baseline CL loss. The classification loss term CL is utilized throughout all experiments, since it is beneficial to the overall performance of the teacher model, as has also been indicated in several related studies in the literature (Hinton et al., 2015; Theodorakopoulos et al., 2020, 2021). Table 5 includes the experimental results (EER with user threshold) for the three offline signature datasets following the Writer Dependent (WD) evaluation with the trained ResNet-18 models for feature extraction. In order to provide a baseline, in the same Table we also report the results obtained by the ResNet-18 model with randomly initialized weights using Xavier initialization (Glorot & Bengio, 2010)). For completeness, the model trained only with the classification objective (CL loss only) is also presented in the Table 5.

Table 5: Performance of the WD classifiers with user threshold on the SigNet-to-ResNet FKD schemes.

| Method | Overall loss | | | | EER (user threshold) | | |
|---|---|---|---|---|---|---|---|
| | $l_1$ | $l_2$ | $g$ | $c$ | GPDS300GRAY ($N_{REF}$=12) | MCYT75 ($N_{REF}$=10) | CEDAR ($N_{REF}$=12) |
| Teacher (fixed) | - | - | - | - | 3.29±0.24 | 3.14±0.60 | 4.33±0.66 |
| Student (RW) | - | - | - | - | 8.57±0.31 | 10.44±0.99 | 9.86±1.31 |
| CL (w/o KD) | 0 | 0 | 0 | 1 | 3.68±0.27 | 3.98±0.75 | 2.39±0.36 |
| CL + KD: GEOM | 10 | 100 | 0 | 1 | 2.95±0.24 | 3.67±0.73 | 2.24±0.45 |
| CL + KD: T-CE | 0 | 0 | 0.001 | 1 | 3.44±0.34 | 4.53±1.04 | 1.92±0.32 |
| CL + KD: BT | 0 | 0 | 0.0001 | 1 | 3.65±0.28 | 4.25±0.77 | 2.37±0.38 |
| CL + KD: BC | 0 | 0 | 0.0001 | 1 | 2.89±0.28 | 3.22±0.63 | 2.07±0.43 |
| CL + KD: GEOM & T-CE | 10 | 100 | 0.001 | 1 | 2.87±0.20 | 3.52±0.74 | 2.19±0.38 |
| CL + KD: GEOM & BT | 10 | 100 | 0.0001 | 1 | 2.97±0.29 | 4.31±0.94 | 1.85±0.32 |
| CL + KD: GEOM & BC | 10 | 100 | 0.0001 | 1 | 2.74±0.28 | 3.29±0.62 | 2.25±0.24 |

### *Comparison between KD and CL losses*

As can be easily inferred from the results, the exploitation of any FKD method together with the CL is advantageous for the student's performance, since the combined optimization of any of the KD terms along with CL is better than CL alone. Also, the combination of local and global KD along with the classification task is the most effective KD method (i.e., CL + KD: GEOM & RESP) considering the performance on all the three datasets. Furthermore, the settings where global KD is realized via our proposed adaptation of Barlow Colleagues (BC) loss achieves the best overall performance for the three signature databases, while Barlow Twins (BT) loss or temperature scaling (T-CE) loss exhibit better results in only one dataset, while degrading results in the others (e.g., CL + KD: T-CE or CL + KD: GEOM & BT).

These conclusions are verified with statistical tests (Friedman and Wilcoxon with p-value at 0.05 on the ten repetitions of classifiers' EERs), comparing the results of S-T training against the sole classification training. In this manner, the KD with BC either in combination with geometrical loss (CL + KD: GEOM & BC) or alone (CL + KD: BC) results to statistically significant improvement of EER on all three datasets while the geometrical regularization alone (CL + KD: GEOM) demonstrates significant difference only on GPDS dataset. The other two response regularization methods (CL + KD: BT, CL + KD: T-CE) as well as their combination with geometric regularization (CL + KD: GEOM & BT, CL + KD: GEOM & T-CE) produce results which are statistically equivalent to those achieved by using only CL loss. Finally, it is interesting to observe that the ResNet model trained only with CL loss, has better results than the SigNet architecture from the previous experiment (Table 4), proving the greater capability of ResNet architecture and confirming the need to utilize more contemporary architectures for OffSV. Weak evidence on that can also be derived from the comparison of the two architectures with random weights, where the ResNet (RW) is superior to SigNet (RW). Ultimately, the S-T scheme provides feature extraction models superior to those obtained by training only to the task of text classification, yet utilizing the same data sources but inducting the prior knowledge of teacher on the OffSV task.

## Comparison between KD and Teacher's performance

The student model resulted from S-T training with FKD via geometrical and response regularizations together with the CL loss, clearly outperforms the teacher in both GPDS and CEDAR datasets. Since the best results obtained when utilizing both local and global based KD, the teacher model is initially compared with these three student models. An in previous, Friedman's test and Wilcoxon paired signed-rank test were used again, with a 5% level of significance, for the ten repetitions of classifiers, using the same permutations of reference and test signatures for the comparisons. The exploitation of geometrical and global KD along with the CL (i.e., CL + KD: GEOM & RESP (of BC, BT, or T-CE)) achieves statistically better performance than the teacher SigNet model for the GPDS and CEDAR datasets while delivering statistical equivalent results in two out of the three cases for the MCYT dataset. For example, the CL + KD: GEOM & BT combination has a bad effect in the performance that can be justified from the different CNN architectures between student and teacher, similar to the comments on the proof-of-concept section above. For completeness, the teacher's performance is also compared to the each of the KD versions individually. The student exhibits statistical difference in performance for all the cases expect that of temperature scaling loss (CL + KD: T-CE) in the GPDS, and the Barlow Twins loss (CL + KD: BT) in the MCYT dataset. Thus, we can observe that the most efficient single KD schemes are those utilizing geometrical loss (CL + KD: GEOM) and BC loss (CL + KD: BC), where the EER values are either lower or not statistically different than those of the teacher. Finally, the ResNet-18 model trained only with CL loss (without KD) is statistically inferior to the teacher model for the GPDS and MCYT datasets and statistically superior to the teacher for the CEDAR dataset. For the case of CEDAR, it is notable that the teacher has degraded performance anyway, probably due to the large canvas size since used universally for all three datasets, in an aim to eliminate unrealistic dataset-dependent pre-processing parameters. At last, the S-T training and specifically the setting with CL + KD: GEOM & BC losses outperforms the teacher on the OffSV problem, without using any signature images for training the feature extraction model.

*Comparison between KD methods*

According to the above, FKD methods using geometrical loss and/or BC response loss are the most beneficial for training feature extraction models. In this section we compare the different KD methods using additional statistical tests to characterize the differences among them. Seven different S-T KD versions in the three signature datasets are compared, and an overview of the findings is provided. The geometrical loss (CL + KD: GEOM) is statistically in tie with the BC response loss (CL + KD: BC) as well as with the two combinations of geometrical and response BT or T-CE losses (CL + KD: GEOM & BT/T-CE) for all the three datasets. Besides, the combination of geometrical and BC losses is statistically superior in the GPDS dataset and statistically equivalent to other two datasets as compared to using solely geometrical loss. Between the response losses, the BC approach is statistically better than the other two losses in GPDS and MCYT, while it has statistically equivalent results with BT and worse than T-CE in CEDAR. Furthermore, the utilization of BC in the combination of local and global KD has statistically better performance in the GPDS and MCYT datasets as well as statistically equivalent results to T-CE and worse behavior than BT in CEDAR dataset. As a general conclusion, the local information seems to have significant importance for the final performance, and also the exploitation of BC regularization in the overall loss reflects a safe and efficient solution across datasets. The regularization of local features on earlier layers guides the training to a higher degree, avoiding the divergence of learning process such could be induced by KD methods relying only on global information where the regularization is applied deeper in the network. Nevertheless, an appropriate response regularization loss could conflate global and local information and capitalize on the joint power of local and global features, as in the case of the BC loss that cooperates efficiently with the geometrical regularization allowing multiple degrees of freedom during the student's training. After all, the greater performance of student over teacher in two out of three datasets (GPDS and CEDAR) and the tantamount of student and teacher results in the other dataset (MCYT) confirms the efficiency of the proposed S-T framework. Also, our choice to utilize a modern ResNet-based architecture (changing from AlexNet-based) has a good impact in order to exploit optimal the knowledge of the teacher. Ultimately, the proposed FKD methods enable the expert CNN (teacher) in signature signals to supervise the learning of the ResNet student without the need to utilize signatures during training and finally provide an effective CNN-based feature extractor for OffSV.

### 4.4.3 Summary of state-of-the-art WD OffSV systems

In this section we to summarize the state-of-the-art (SoTA) methods for Writer Dependent (WD) OffSV systems, evaluating their performance to the proposed system. Given the inherent differences between various methods, there are many additional variables in the implementation of the systems' stages that renders the task of fairly comparing all of them very difficult. Hence, the purpose of the presentation of SoTA literature is to provide a general overview for the WD OffSV field, denoting the most important results in the three most popular datasets of CEDAR, MCYT75, and GPDS-300-GRAY. Table 6 presents a summary of the related SoTA works for WD OffSV task using the EER metric. Also, it includes the number of reference signatures used to form the positive class for training the WD classifiers. A common number of reference signatures could be found across methods for each dataset, despite the differences between methods as well as the different approaches for selecting these reference signatures.

Table 6: Summary of state-of-the-art OffSV Systems in terms of EER metric, for the CEDAR, MCYT75, and GPDS300GRAY datasets.

| Refs | Method | GPDS300GRAY | | MCYT75 | | CEDAR | |
|---|---|---|---|---|---|---|---|
| | | $N_{REF}$ | EER | $N_{REF}$ | EER | $N_{REF}$ | EER |
| (Soleimani et al., 2016) | HOG + DMML | 10 | 20.94 | 10 | 9.86 | - | - |
| (Serdouk et al., 2017) | HOT | 12 | 9.30 | 10 | 10.60 | - | - |
| (Diaz et al., 2017) | Duplicator | 12 | 14.58 | 12 | 9.12 | - | - |
| (Hafemann et al., 2017a) | SigNet | 12 | 3.15 | 10 | 2.87 | 12 | 4.76 |
| (Hafemann et al., 2017a) | SigNet-F | 12 | 1.69 | 10 | 3.00 | 12 | 4.63 |
| (Hafemann et al., 2018) | SigNet-SPP | 12 | 0.41 | 10 | 3.64 | 10 | 3.60 |
| (Lai & Jin, 2018) | PDSN | - | - | 10 | 3.78 | 10 | 4.37 |
| (Zois et al., 2019) | SR – KSVD/OMP | 12 | 0.70 | 10 | 1.37 | 10 | 0.79 |
| (Bhunia et al., 2019) | Hybrid Texture | 12 | 8.03 | 10 | 9.26 | 10 | 6.66 |
| (Maergner et al., 2019) | CNN-Triplet and Graph edit distance | - | - | 10 | 3.91 | 10 | 5.91 |
| (Shariatmadari et al., 2019) | HOCCNN | - | - | 12 | 5.46 | 12 | 4.94 |
| (Mersa et al., 2019) | ResNet trained with text | - | - | 10 | 3.98 | - | - |
| (Masoudnia et al., 2019) | MLSE | - | - | 10 | 2.93 | - | - |
| (Zois et al., 2020) | Visibility Motif profiles | - | - | 10 | 1.54 | 10 | 0.51 |
| (Maruyama et al., 2021) | SigNet-F classifier gauss augments | 3 | 0.20 | 3 | 0.01 | 3 | 0.82 |
| (Liu et al., 2021) | MSDN | - | - | - | - | 10 | 1.75 |
| (Yapıcı et al., 2021) | Cycle-GAN | - | - | 10 | 2.58 | - | - |
| (Zheng et al., 2021) | micro deformations | - | - | - | - | 12 | 2.76 |
| (Tsourounis et al., 2022) | CNN-CoLL | 12 | 2.12 | 10 | 1.62 | 10 | 1.66 |
| (Viana et al., 2023) | MT-SigNet (Triplet) | - | - | 10 | 2.71 | 12 | 3.50 |
| (Viana et al., 2023) | MT-SigNet (NT-Xent) | - | - | 10 | 3.22 | 12 | 3.32 |
| Proposed | S-T FKD (CL + KD: GEOM & BC) | 12 | 2.74 | 10 | 3.29 | 12 | 2.25 |

The OffSV systems consist of three stages: the preprocessing, the feature extraction, and the classifier. These stages are designed according to the characteristics of each method and thus, many influential technicalities exist, like different preprocessing steps (e.g., such as different data preparation procedures of (Hafemann et al., 2017a; Zois et al., 2019), different input image size (Liu et al., 2021; Zheng et al., 2021), etc.), types of classifiers (e.g., like using SVM (Hafemann et al., 2017a; Masoudnia et al., 2019; Serdouk et al., 2017; Viana et al., 2023; Zois et al., 2019), one-class SVM (Bhunia et al., 2019; Lai & Jin, 2018)[Lai2018], [Bhunia2019], Artificial Neural Networks (ANN) (Shariatmadari et al., 2019; Yapıcı et al., 2021), thresholding (Maergner et al., 2019; Soleimani et al., 2016), etc.) as well as major differences such as the type of training data (e.g., different training signature datasets (Bhunia et al., 2019; Hafemann et al., 2017a), private dataset (Liu et al., 2021), auxiliary data (Mersa et al., 2019; Tsourounis et al., 2022), augmentation or synthetic data (Diaz et al., 2017; Maruyama et al., 2021; Viana et al., 2023; Yapıcı et al., 2021), etc.). Although we have chosen so that the presented OffSV systems do not utilize skilled forgery signatures in the classifier's training and the number of reference signatures be common in many cases, the varying amount of the negative training class at the classifier has a major impact in the performance too (as also reported in the works of (Hafemann et al., 2017a) and (Maruyama et al., 2021)). Additionally, the number of test samples differs since some methods utilize all the available signatures, considering the rest of the genuine and all the skilled forgeries (e.g., (Zois et al., 2019)), while other methods use equal number of test genuine and skilled forgery signatures -by selecting randomly the test skilled forgeries- (e.g., our implementation,(Hafemann et al., 2017a; Maruyama et al., 2021; Viana et al., 2023)). Hence, easy comparisons between methods could be misleading and just a general outlook should be extracted. In this manner, we could argue that the proposed OffSV system proves the feasibility of achieving a low verification error, which is at least comparable to the state-of-the-art methods in all three datasets, despite nor using any signatures for training the feature extraction model.

### 4.4.4 Comparisons with SigNets

Finally, in this section we extensively compare the obtained models with two variants of SigNet: the teacher SigNet model and the SigNet-F model that used both genuine and skilled forgery signatures during its training (with the now defunct GPDS960 corpus) (Hafemann et al., 2017a). The comparisons are performed following identical preprocessing steps and utilizing the same partition of training and test signatures for the evaluated models to cast the comparisons as fair as possible. Thus, the respective classifiers are developed with common training and test sets across compared models. In this manner, the comparisons focus on the performance of the feature extraction stage and reveal the models' efficiency. Additionally, the calculation of multiple metrics provides a detailed performance analysis. Table 7 presents the performance derived by the three compared feature extraction models: the SigNet-F, the SigNet (teacher) and our proposed model (ResNet from S-T KD with CL + KD: GEOM & BC) in the three signature datasets. The performance is evaluated using five different metrics, the False Rejection Rate (FRR), the False Acceptance Rate (FAR) on skilled forged signatures, the EER values when global and user-specific thresholds are utilized, and the mean Area Under Curve (AUC) using the Receiver Operating Characteristic (ROC) curves. Also, the number of reference signatures ($N_{REF}$) ranges from 3, 5, 10, up to 12 signatures.

Table 7: Detailed comparison with SigNet and SigNet-F. All the reported metrics were obtained using WD classifiers while the FRR and FAR-skilled metrics are measured when the threshold is set to zero.

| Dataset | Method | N<sub>REF</sub> | FRR | FAR skilled | EER global threshold | EER user thresholds | AUC |
|---|---|---|---|---|---|---|---|
| GPDS 300GRAY | SigNet-F | 3 | 6.26 (± 0.26) | 6.25 (± 0.27) | 6.25 (± 0.26) | 2.61 (± 0.30) | 99.10 (± 0.12) |
| | | 5 | 5.01 (± 0.16) | 5.01 (± 0.17) | 5.01 (± 0.16) | 2.04 (± 0.21) | 99.37 (± 0.08) |
| | | 10 | 4.06 (± 0.16) | 4.06 (± 0.17) | 4.06 (± 0.16) | 1.68 (± 0.08) | 99.55 (± 0.05) |
| | | 12 | 3.93 (± 0.16) | 3.92 (± 0.16) | 3.92 (± 0.16) | 1.53 (± 0.15) | 99.57 (± 0.06) |
| | SigNet (Teacher) | 3 | 9.29 (± 0.25) | 9.29 (± 0.24) | 9.29 (± 0.24) | 4.79 (± 0.31) | 97.92 (± 0.14) |
| | | 5 | 7.81 (± 0.24) | 7.79 (± 0.25) | 7.80 (± 0.25) | 4.12 (± 0.26) | 98.32 (± 0.12) |
| | | 10 | 6.28 (± 0.19) | 6.27 (± 0.18) | 6.27 (± 0.18) | 3.42 (± 0.21) | 98.65 (± 0.13) |
| | | 12 | 6.02 (± 0.30) | 6.05 (± 0.30) | 6.04 (± 0.30) | 3.29 (± 0.24) | 98.72 (± 0.08) |
| | ResNet S-T FKD (Student) | 3 | 8.76 (± 0.30) | 8.77 (± 0.31) | 8.76 (± 0.30) | 4.76 (± 0.31) | 98.05 (± 0.16) |
| | | 5 | 7.33 (± 0.23) | 7.34 (± 0.24) | 7.33 (± 0.23) | 3.76 (± 0.33) | 98.55 (± 0.14) |
| | | 10 | 5.49 (± 0.17) | 5.51 (± 0.16) | 5.50 (± 0.16) | 2.86 (± 0.23) | 98.93 (± 0.13) |
| | | 12 | 5.23 (± 0.23) | 5.26 (± 0.20) | 5.25 (± 0.22) | 2.74 (± 0.28) | 99.01 (± 0.10) |
| MCYT75 | SigNet-F | 3 | 10.05 (± 0.45) | 10.09 (± 0.47) | 10.07 ± 0.44) | 5.99 (± 0.69) | 97.16 (± 0.54) |
| | | 5 | 7.36 (± 0.68) | 7.40 (± 0.65) | 7.38 (± 0.66) | 3.77 (± 0.71) | 98.32 (± 0.39) |
| | | 10 | 6.32 (± 0.55) | 6.30 (± 0.55) | 6.31 (± 0.54) | 3.19 (± 0.52) | 98.52 (± 0.33) |
| | | 12 | 5.42 (± 0.52) | 5.48 (± 0.69) | 5.45 (± 0.58) | 2.20 (± 0.58) | 98.95 (± 0.35) |
| | SigNet (Teacher) | 3 | 9.49 (± 0.77) | 9.46 (± 0.75) | 9.48 (± 0.76) | 4.79 (± 0.87) | 97.68 (± 0.45) |
| | | 5 | 7.15 (± 0.75) | 7.05 (± 0.79) | 7.10 (± 0.76) | 3.86 (± 0.74) | 98.21 (± 0.54) |
| | | 10 | 6.51 (± 0.40) | 6.35 (± 0.38) | 6.43 (± 0.38) | 3.14 (± 0.60) | 98.69 (± 0.21) |
| | | 12 | 6.09 (± 0.63) | 6.19 (± 0.64) | 6.14 (± 0.62) | 2.73 (± 0.80) | 98.80 (± 0.35) |
| | ResNet S-T FKD (Student) | 3 | 13.31 (± 1.21) | 13.32 (± 1.29) | 13.32 (± 1.24) | 7.94 (± 1.24) | 94.70 (± 1.18) |
| | | 5 | 10.51 (± 0.46) | 10.55 (± 0.46) | 10.53 (± 0.46) | 5.84 (± 0.86) | 96.18 (± 1.08) |
| | | 10 | 7.12 (± 0.57) | 6.03 (± 0.34) | 6.49 (± 0.34) | 3.29 (± 0.62) | 98.43 (± 0.25) |
| | | 12 | 6.84 (± 0.92) | 7.01 (± 1.07) | 6.93 (± 0.99) | 3.13 (± 0.66) | 98.03 (± 0.52) |
| CEDAR | SigNet-F | 3 | 16.29 (± 0.68) | 16.29 (± 0.79) | 16.29 (± 0.73) | 9.52 (± 0.95) | 93.92 (± 0.88) |
| | | 5 | 14.02 (± 1.00) | 14.13 (± 1.04) | 14.07 (± 1.02) | 8.78 (± 1.05) | 94.76 (± 1.00) |
| | | 10 | 10.80 (± 0.78) | 10.84 (± 0.91) | 10.82 (± 0.85) | 6.54 (± 0.97) | 96.19 (± 0.67) |
| | | 12 | 10.60 (± 0.59) | 10.60 (± 0.54) | 10.60 (± 0.56) | 5.99 (± 0.64) | 96.66 (± 0.49) |
| | SigNet (Teacher) | 3 | 13.51 (± 0.71) | 13.47 (± 0.79) | 13.49 (± 0.75) | 6.75 (± 1.23) | 96.27 (± 0.79) |
| | | 5 | 11.22 (± 0.68) | 11.18 (± 0.71) | 11.20 (± 0.68) | 5.92 (± 0.47) | 97.04 (± 0.28) |
| | | 10 | 8.33 (± 0.37) | 8.36 (± 0.41) | 8.35 (± 0.38) | 4.34 (± 0.72) | 97.84 (± 0.27) |
| | | 12 | 7.98 (± 0.55) | 7.98 (± 0.58) | 7.98 (± 0.56) | 4.33 (± 0.66) | 97.84 (± 0.39) |
| | ResNet S-T FKD (Student) | 3 | 7.85 (± 1.06) | 7.85 (± 1.05) | 7.85 (± 1.05) | 3.39 (± 0.47) | 98.50 (± 0.41) |
| | | 5 | 6.35 (± 0.90) | 6.24 (± 0.81) | 6.29 (± 0.85) | 2.85 (± 0.27) | 98.70 (± 0.29) |
| | | 10 | 4.78 (± 0.43) | 4.71 (± 0.36) | 4.75 (± 0.39) | 2.20 (± 0.33) | 99.11 (± 0.16) |
| | | 12 | 4.16 (± 0.36) | 4.13 (± 0.37) | 4.15 (± 0.36) | 2.25 (± 0.24) | 99.10 (± 0.19) |

For each setting, Table 7 provides different qualities of the system's performance across the horizontal direction via the five calculated metrics while the varying number of reference signatures provides a different view of classifier's effectiveness. In this manner, we could derive a few important observations. First, the number of reference signatures has a critical impact on the performance of an OffSV system since decreasing the reference samples causes shrinkage on both the positive and the negative training class of the SVM. Hence, only a robust and discriminative feature extractor could assist the classifier to address the problem with a small amount of training samples. Secondly, the SigNet is clearly better than SigNet-F in CEDAR dataset and worse than SigNet-F in GPDS dataset. The inferior performance of SigNet in GPDS dataset though, is something totally reasonable because the knowledge of forged signatures from the same dataset is implicitly encoded in the SigNet-F offering a performance edge, given the later was trained with both genuine and skilled forgeries, even they originated from different writers of GPDS960 corpus (note that GPDS300 is a subset of GPDS960). For the MCYT dataset, the reported results are more complicated since the OffSV system using SigNet exhibits better performance on some metrics (e.g., global threshold for 3 or 5 references) and on the system based on SigNet-F on some other metrics (e.g., user-specific threshold for 3 references), while both systems are equivalent when 10 reference signatures are utilized. In the case of 12 reference signatures, only 3 genuine signatures remain for testing and consequently, the evaluation depends mainly on the test skilled forgery samples that is a less reliable indicator of performance (but it is included in the table for consistency with the other two datasets). In this manner, the performance disparity in MCYT dataset, given the amount of reference samples, relies on the classifier's effectiveness too, meaning that the SVM demands more training signatures to generalize well for all the cases. Ultimately, based on the above observations we can conclude that the SigNet exhibits greater generalization ability than SigNet-F across datasets and is actually a better choice as a single teacher in a S-T scheme.

In the light of the above, we finally compare the performance of ResNet to that of SigNet and SigNet-F on all datasets. The ResNet model trained with the proposed scheme is statistically superior to SigNet and statistically inferior to SigNet-F for the GPDS dataset for the same reasons mentioned above, while ResNet's higher performance is unambiguous for the CEDAR dataset. Thus, the student model achieves to surpass the teacher in both evaluated datasets. In fact, the utilization of a supplementary teacher, like SigNet-F, in a multiple-teachers FKD scheme maybe has positive impact and opens an interesting new path, but it is out of the scope of the current work. Lastly for the MCYT dataset, the ResNet has statistically equivalent performance with SigNet and SigNet-F for 10 reference signatures, whereas the system based on ResNet has inferior performance if the classifier needs to be trained with 3 or 5 genuine samples for the positive class. In conclusion, the experimental evidence from the three datasets support that the proposed FKD scheme comprises an adequate solution to transfer the knowledge from an expert CNN in the field of OffSV into a new (and modern) architecture, without the need fof any signature images, thus helping to create a new generation of models that could offer an excellent initial baseline for further research in Deep-Learning techniques for the OffSV problem.

# 5 Conclusions

In this work, we proposed a Feature-based Knowledge Distillation (FKD) learning framework applied to the OffSV problem. In the presented Student-Teacher (S-T) learning configuration, the knowledge is transferred from a benchmark CNN that provides efficient feature representations for signature images (acting as the teacher) into a new CNN model of different architecture (acting as the student). The only compatibility requirement between the teacher and student models are the spatial matching for at least some of intermediate activations and the common global feature dimensionality. We distilled knowledge through multiple layers via multiple connections among student and teacher topologies in order to incorporate both local and global information. We expressed the local information utilizing a manifold-to-manifold distance function that is designed to match the manifolds of local activations at the different layers of teacher and student models through geometric criteria of dissimilarity. Additionally, we promoted similarity in the feature responses of the student and teacher models by using loss functions that incorporate temperature-scaled cross-entropy or normalized cross-correlation to force the student's global features to imitate those of the teacher. The latter approach also included a novel loss function that leverages the cross-correlation matrix between the global features extracted from the student and teacher models respectively, considering the different architectures between the two CNNs. Hence, we presented, for the first time, a solution to inherit the prior knowledge of an effective CNN model into a new CNN model for signature representation learning through a KD method.

Since we did not have access to the signatures that the teacher model is trained on, we used auxiliary data from a related domain, such as images of handwritten text, as a source of information for KD. In this manner, we take advantage from the resemblance between handwriting in both texts and signatures, to overcome the lack of large amount of signature images that are required for the S-T training. The proposed S-T FKD framework is strengthened by utilizing knowledge from multiple layers and advanced relation-based distillation algorithms. These capture the correlations and higher-order output dependencies between the teacher and student models, enabling the student model to acquire the knowledge from a fixed teacher to a very large extent. Hence, when the response loss, calculated using the proposed Barlow-Colleagues (BC) cross-correlation function, is combined with the geometric loss, based on manifold-to-manifold distance, the student model becomes at least as efficient as the teacher, if not more so.

A significant motivation for this work was to enable the use of modern deep-learning models in OffSV, despite the current unavailability of a large signature dataset. Our FKD scheme serves as an excellent starting point for further research, as it incorporates knowledge from efficient OffSV models and addresses the lack of publicly available signature datasets. The presented OffSV system also shares common pre-processing and decision stages with other state-of-the-art methods, allowing for fair comparisons and emphasizing the unique contributions of each approach.

However, there are certain limitations in our study. The dependence on specific pre-processing steps for both signatures and text images restricts the flexibility of the system, as the teacher model has learned to encode information from pre-processed signatures. Nevertheless, we tried to mitigate the impact using common parameters during the signature pre-processing for all the evaluating datasets. We did not extensively analyze scale variations between raw text and signature images, limiting our understanding in that aspect since we addressed this implicitly by applying different canvas sizes during the generation of the training text images. Additionally, designing the student architecture with (spatially) matching intermediate activations and global feature dimension consistency with the teacher model, could pose some challenges for some architectures which may be a limiting factor from an architectural design perspective.

Ultimately, the main contribution of this work is the efficient knowledge transfer from a teacher model to any new architecture, as demonstrated with the use of a ResNet student. This allows for the leveraging of the efficiency of a basic and outdated teacher architecture, and the transfer of that knowledge into a deeper and more advanced student architecture for the efficient encoding of signatures. As a result, the student CNN can provide efficient feature representations for the OffSV task, even without utilizing any signature images during the FKD process.

Future work will include investigation of additional knowledge distillation methods and different distillation strategies. We believe that combining multiple teachers and multi-loss approaches could be promising for the OffSV task since they could improve the generalization ability of the student model and make it more effective across signature datasets and languages. Additionally, future plans could incorporate synthetic signatures, using both existing synthetic datasets and generative methods to generate new signature images, during training a KD scheme. However, the main challenge in this research direction is the difference between the distributions of original and synthetic signatures, resulting in models that are only efficient on one or the other type of data. Another useful research approach is to develop CNN-based schemes that leverage diverse pre-processing steps, allowing for the decoupling of specialized preprocessing techniques tailored to specific datasets. This approach can be further enhanced by incorporating synthetic samples to augment the training process, particularly where a large number of real signatures is not available. All the aforementioned approaches could be also organized on a comprehensive study that involves a variety of KD methods for the OffSV problem, including both WI and WD evaluation phases for optimal deliberation. Finally, including few-shot learning techniques into a S-T framework would be an interesting area of future research.

**Authors contribution**

**Dimitrios Tsourounis:** Conceptualization, Methodology, Software, Investigation, Formal analysis, Writing – original draft, Writing – review & editing. **Ilias Theodorakopoulos:** Conceptualization, Methodology, Validation, Supervision, Writing – review & editing. **Elias N. Zois:** Visualization, Supervision. **George Economou:** Resources, Supervision.

**Declaration of competing interest**

The authors declare that they have no known competing financial interests or personal relationships that could have appeared to influence the work reported in this paper.

**Funding**

This research is co-financed by Greece and the European Union (European Social Fund- ESF) through the Operational Program «Human Resources Development, Education and Lifelong Learning» in the context of the project "Strengthening Human Resources Research Potential via Doctorate Research – 2nd Cycle" (MIS-5000432), implemented by the State Scholarships Foundation (IKY).

https://openaccess.thecvf.com/content_cvpr_2017/html/Xie_Aggregated_Residual_Transformations_CVPR_2017_paper.html

Xiong, Y.-J., & Cheng, S.-Y. (2021). Attention Based Multiple Siamese Network for Offline Signature Verification. In J. Lladós, D. Lopresti, & S. Uchida (Eds.), *Document Analysis and Recognition – ICDAR 2021* (pp. 337–349). Springer International Publishing. https://doi.org/10.1007/978-3-030-86334-0_22

Yapıcı, M. M., Tekerek, A., & Topaloğlu, N. (2021). Deep learning-based data augmentation method and signature verification system for offline handwritten signature. *Pattern Analysis and Applications*, *24*(1), 165–179. https://doi.org/10.1007/s10044-020-00912-6

Yilmaz, M. B., & Öztürk, K. (2018). Hybrid User-Independent and User-Dependent Offline Signature Verification with a Two-Channel CNN. *2018 IEEE/CVF Conference on Computer Vision and Pattern Recognition Workshops (CVPRW)*, 639–6398. https://doi.org/10.1109/CVPRW.2018.00094

Yılmaz, M. B., & Öztürk, K. (2020). Recurrent Binary Patterns and CNNs for Offline Signature Verification. In K. Arai, R. Bhatia, & S. Kapoor (Eds.), *Proceedings of the Future Technologies Conference (FTC) 2019* (pp. 417–434). Springer International Publishing. https://doi.org/10.1007/978-3-030-32523-7_29

Yonekura, D. C., & Guedes, E. B. (2021). Offline Handwritten Signature Authentication with Conditional Deep Convolutional Generative Adversarial Networks. *Anais do Encontro Nacional de Inteligência Artificial e Computacional (ENIAC)*, 482–491. https://doi.org/10.5753/eniac.2021.18277

Younesian, T., Masoudnia, S., Hosseini, R., & Araabi, B. N. (2019). Active Transfer Learning for Persian Offline Signature Verification. *2019 4th International Conference on Pattern Recognition and Image Analysis (IPRIA)*, 234–239. https://doi.org/10.1109/PRIA.2019.8786013

Zbontar, J., Jing, L., Misra, I., LeCun, Y., & Deny, S. (2021). Barlow Twins: Self-Supervised Learning via Redundancy Reduction. *Proceedings of the 38th International Conference on Machine Learning*, 12310–12320. https://proceedings.mlr.press/v139/zbontar21a.html

Zemel, R., & Carreira-Perpiñán, M. (2004). Proximity Graphs for Clustering and Manifold Learning. *Advances in Neural Information Processing Systems*, *17*. https://proceedings.neurips.cc/paper/2004/hash/dcda54e29207294d8e7e1b537338b1c0-Abstract.html

Zheng, Y., Iwana, B. K., Malik, M. I., Ahmed, S., Ohyama, W., & Uchida, S. (2021). Learning the Micro Deformations by Max-pooling for Offline Signature Verification. *Pattern Recognition*, 108008. https://doi.org/10.1016/j.patcog.2021.108008

Zhu, Y., Lai, S., Li, Z., & Jin, L. (2020). Point-to-Set Similarity Based Deep Metric Learning for Offline Signature Verification. *2020 17th International Conference on Frontiers in Handwriting Recognition (ICFHR)*, 282–287. https://doi.org/10.1109/ICFHR2020.2020.00059

Zois, E. N., Said, S., Tsourounis, D., & Alexandridis, A. (2023). Subscripto multiplex: A Riemannian symmetric positive definite strategy for offline signature verification. *Pattern Recognition Letters*, *167*, 67–74. https://doi.org/10.1016/j.patrec.2023.02.002

Zois, E. N., Tsourounis, D., Theodorakopoulos, I., Kesidis, A. L., & Economou, G. (2019). A Comprehensive Study of Sparse Representation Techniques for Offline Signature Verification. *IEEE Transactions*
40